\documentclass[review]{elsarticle}

\usepackage{multirow}
\usepackage{color}
\usepackage{amsmath,amssymb}
\usepackage[ruled,vlined]{algorithm2e}
\usepackage{lineno,hyperref}
\modulolinenumbers[5]

\journal{Journal of Pattern Recognition}

   \graphicspath{{Figures/}}
   \DeclareGraphicsExtensions{.eps,.pdf,.jpg,.png,.gif}

\begin{document}

\begin{frontmatter}

 \title{Online Learnable Keyframe Extraction in Videos and its Application with Semantic Word Vector in Action Recognition}





\author[mymainaddress1]{G M Mashrur~E~Elahi \corref{mycorrespondingauthor}}
\cortext[mycorrespondingauthor]{Corresponding author}
\ead{eelahi@ualberta.ca}

\author[mymainaddress1]{Yee-Hong~Yang}
\ead{herberty@ualberta.ca}

\address[mymainaddress1]{Departments~of~Computing~Science, University of Alberta, T6G 2E8, Canada}

\begin{abstract}
Video processing has become a popular research direction in computer vision due to its various applications such as video summarization, action recognition, etc. Recently, deep learning-based methods have achieved impressive results in action recognition. However, these methods need to process a full video sequence to recognize the action, even though most of these frames are similar and non-essential to recognizing a particular action. Additionally, these non-essential frames increase the computational cost and can confuse a method in action recognition. Instead, the important frames called keyframes not only are helpful in the recognition of an action but also can reduce the processing time of each video sequence for classification or in other applications, e.g. summarization. As well, current methods in video processing have not yet been demonstrated in an online fashion. 
Motivated by the above, we propose an online learnable module for keyframe extraction. 
This module can be used to select key-shots in video and thus can be applied to video summarization. 
The extracted keyframes can be used as input to any deep learning-based classification model to recognize action. We also propose a plugin module to use the semantic word vector as input along with keyframes and a novel train/test strategy for the classification models. 
To our best knowledge, this is the first time such an online module and train/test strategy  have been proposed.
The experimental results on many commonly used datasets in video summarization and in action recognition have shown impressive results using the proposed module.

\end{abstract}


\end{frontmatter}


\section{Introduction}
     Nowadays, the availability of a huge volume of video has attracted the attention of many researchers in the field of video analysis such as video summarization and action recognition \cite{PR1_hussain109comprehensive,Unpaired:rochan2019video,S1_weinland2011survey}. 
     In video summarization, the objective is to represent a  video sequence using representative shots, or key shots \cite{PR2_wu2020dynamic,Susinet:koutras2019susinet} while in action recognition, the objective is to classify a video sequence \cite{S2_poppe2010survey,S1_weinland2011survey}. 
     Human-action recognition from videos is getting more popular due to its increasing role in many applications from surveillance to autonomous driving \cite{MLDA_yan2014multitask,HFS_zhu2016handcrafted}. Recent action recognition methods rely mostly on deep learning because of their impressive classification accuracy
    \cite{carreira2017quo,STM:jiang2019stm}.
    In deep learning based action recognition, the most important features are the spatiotemporal features \cite{STM:jiang2019stm}, which encode the relationship of spatial features from different timestamps and motion features between neighboring frames.
    
   The existing methods of video understanding using deep learning can be categorized into two-stream neural networks \cite{STM:jiang2019stm,TwoStream4:wang2017spatiotemporal} and 3D convolution networks (3D CNNs) \cite{3Dconv1:qiu2017learning,LCDC:mac2019learning}. The two-stream network has an RGB stream with RGB frames and a flow stream with optical flow as input. However, the flow stream in such networks lacks the ability to capture long term temporal relationship \cite{STM:jiang2019stm}. Moreover, the extraction of optical flow is highly expensive in both time and space which restricts using such methods in real world applications. 
   Later, 3D CNN based methods have been proposed which can represent the temporal and spatial features together and overcome some of the limitations of the two-stream methods \cite{3Dconv1:qiu2017learning}. 3D CNNs can capture long-range temporal relationship using stacked 3D convolutions. Although these methods have achieved superior performance by pre-training on large-scale video datasets, their use of 3D convolution has increased the computational cost enormously compared to that of using 2D convolution. As well, existing 3D methods use a full video sequence to recognize its corresponding action which not only increases the computational cost but also is unnecessary because most of the frames in a video sequence are similar. 
   It would be best if the methods could extract important frames in a video sequence that would be sufficient to represent the video and could be used to recognize the corresponding action. This is, indeed, the motivation of our proposed method.  
   
   Recently, keyframe-based approaches for action recognition have become increasingly popular because of their low computational costs and low storage requirements \cite{PR3_liu2013boosted,MKF_carlsson2001action,key-framing_raptis2013poselet,SFVpeng2014action,KeyVolzhu2016key}. These approaches rely on the intuition that a set of {\em keyframes} can be selected from all the frames in a video sequence to sufficiently represent the action in the video sequence. The concept of keyframe allows a recognition method to focus on the most distinctive part of a sequence and to ignore frames that are not discriminative or relevant \cite{key-framing_raptis2013poselet}. However, the above mentioned approaches suffer from four main limitations: 1) they may rely on manual selection of keyframes \cite{MKF_carlsson2001action}, 2) they may fail to classify actions with large variations \cite{SFVpeng2014action}, 3) they may require a complex learning process to decide which set of frames extract \cite{KeyVolzhu2016key}, and most importantly, 4) they are not online i.e. the whole sequence is required to select keyframes.
   
   In this paper, we propose the first online learnable module to extract keyframes for video representation that addresses the above mentioned limitations. Our proposed module is inspired by the method of Mac, et al. \cite{LCDC:mac2019learning}, which uses locally-consistent deformable convolution to extract spatiotemporal features for action recognition. We leverage the deformable convolution information in our module to extract keyframes. 
   Unlike their method, our proposed module extract keyframes using the motion information of a video sequence. The keyframes are selected based on a pixel-wise learnable threshold ($TH$) along with other convolution parameters. Moreover, the proposed module is an online module as it processes frames in a video sequence one at a time and selects keyframes using the information of immediate neighboring frames. Our module is simple to train and can be used in many applications in computer vision. For example, the extracted keyframes can be used as input to a typical deep learning action recognition method for action recognition or to get key shots for video summarization. 
   
   Keyframes in action recognition and key shots in video summarization have similar objectives. Hence, to demonstrate the effectiveness of our proposed module, we evaluate our results on two standard video summarization datasets: SumMe \cite{SumMe:gygli2014creating} and TVSum \cite{TVSum:song2015tvsum}. Using the proposed keyframe extraction method on video summarization have shown impressive results on the benchmark datasets. 
   In particular, our method achieves the best state-of-the-art results on summarization datasets in the transfer learning environment.
   Most importantly, our method extracts a fewer number of keyframes per sequence that is on average 25\%-30\% of the full video sequence and is 50\% less than that of state-of-the-art keyframe-based action recognition methods which typically require almost 50\% of the video sequence to achieve competitive recognition results.
   
   For action recognition, we use three well-known benchmark action recognition datasets: Kinetics-400 \cite{Kinetic400:kay2017kinetics}, Something-Something-V1 \cite{SSv1dataset:goyal2017something}, and HMDB51 \cite{Dataset_hmdb51}. We
   extract the keyframes from the action recognition datasets using our proposed module trained with video summarization datasets.
   We apply the extracted keyframes as input to the well-known 3D CNN methods to recognize action. Using only keyframes, these methods achieve better results compared to that of using the full video sequence for some datasets and comparable results in other datasets. 
   To further improve the results of action recognition, for the first time we adopt the well-known semantic word vector (W2V) \cite{W2V_pennington2014glove} of the action class labels along with the keyframes of the videos of the corresponding action labels as input to the classification models. Towards this, we propose a plugin module that combines and refines the semantic word vectors in an iterative manner, and use it along with the keyframe features for classification. To achieve this goal we also propose a novel iterative train/test strategy called the ITTS approach for the classification models. 
   Our proposed approach using W2V and keyframes as input achieves even better state-of-the-art results on the three benchmark action recognition datasets.   
   These encouraging results demonstrate the effectiveness of our proposed module of keyframe extraction as well as the proposed approach of using W2V in action recognition.   
   
   The contributions of our work are summarized as follows: 
   \begin{itemize}
       \item[\textbullet] We propose a new online learnable keyframe extraction module to extract keyframes in a video sequence and use a learnable threshold kernel to select keyframes. 
       %
       %
       \item[\textbullet] The proposed module is the first online module that can process frames sequentially and select the keyframes based on processed frames only. This makes the module fast and memory efficient because it does not require to store all the past frames' information. To our best knowledge, this is the first time that such an online module has been proposed. 
       \item[\textbullet] Our online keyframe extraction module can be used for transfer learning as well. In particular, training the model with one dataset, e.g. summarization datasets, and applying the trained model to another dataset, e.g. action recognition datasets, achieves state-of-the-art results in action recognition. 
       Moreover, our method achieves the best state-of-the-art results on summarization datasets in the transfer learning environment.
       \item[\textbullet] We also propose a plugin module and a novel train/test strategy called ITTS for the first time to incorporate the semantic word vector along with keyframe features as input to the classification models. Our proposed approach of action recognition further improves the best state-of-the-art results on the action recognition datasets.
       
       %
   \end{itemize}

\section{Related Works}
    \subsection{Deep Learning Methods} 
        Recently, deep learning based action recognition methods have achieved impressive performance in the classification accuracy of human action recognition \cite{carreira2017quo,32hmdb_wang2015action,STM:jiang2019stm}. However, the models of these methods are computationally expensive to train. Among these methods, the two-stream Inflated 3D ConvNet (I3D) \cite{carreira2017quo} has achieved the best accuracy on the HMDB51 \cite{Dataset_hmdb51} dataset so far. This method is based on the 2D ConvNet which is inflated to 3D by extending the dimension of the filter kernels. The method uses two separate streams, one for the RGB and the other the optical flow input. 
        %
        Another deep network based method employs spatiotemporal convolutions on action recognition \cite{closer_tran2018closer}. The authors introduce (2+1)D convolution which explicitly factorizes 3D convolution into a 2D spatial convolution and a 1D temporal convolution. Experimentally, this is shown to perform better than the direct 3D convolution. Moreover, most of the deep networks also use optical flow to incorporate motion information in action recognition. In most of the two-stream neural networks \cite{STM:jiang2019stm,TwoStream4:wang2017spatiotemporal}, the optical flow information is calculated outside of the network and is the most computationally expensive part of the whole process. 

        Some recent methods have tried to overcome this issue by incorporating the flow calculation using the deep network \cite{image2flow_gao2017im2flow,whatmakes_huang2018makes,optical_sun2018optical,LCDC:mac2019learning}. Instead of using optical flow, the recently proposed optical flow guided feature for action recognition can be efficiently computed using the Sobel operator and used with any CNN architecture  \cite{optical_sun2018optical}. 
        The Locally-Consistent Deformable Convolution (LCDC) method \cite{LCDC:mac2019learning} is proposed to extract spatial and temporal information jointly for action recognition. It tracks the deformation in motion in consecutive frames, from which the spatiotemporal features for each frame are extracted. Inspired by this method, we propose our keyframe extraction module which leverages the LCDC information to extract the keyframes. Instead of using spatiotemporal features for all the frames in a video sequence, we extract the keyframes first and use them to perform action recognition based on the spatiotemporal features extracted from the keyframes.
        Another flow based deep network is proposed where the flow information is hallucinated from a single frame of the video sequence \cite{image2flow_gao2017im2flow}. Interestingly, this method also uses a single frame from a video sequence to recognize the action and its competitive results further confirm the importance of keyframes and motivate us to focus on keyframe based methods. In contrast to flow based methods, quantitative and qualitative analysis of the effect of motion in video action recognition in the context of deep networks is discussed \cite{whatmakes_huang2018makes}. The authors assume that potentially there might be some keyframes that are crucial to recognize action from a video sequence. They experimentally found that a significant number of actions actually do not require motion information in recognition. This method also supports our goal to focus on keyframe based methods. 
        
    \subsection{Keyframe-based deep methods}
        There are some recent keyframe based methods using deep networks that have performed better than conventional feature based methods \cite{SFVpeng2014action}. These methods consider using sub-volumes of a video sequence and use a combination of sub-volumes to extract features for action recognition. One of the promising methods uses the Stacked Fisher Vectors representation of a video \cite{SFVpeng2014action}. In particular, multiple sub-volumes of a video sequence are selected, from which potential sub-volumes are chosen to extract Fisher vectors. Then, the video is represented using multiple layers of nested Fisher vector encoding. However, this approach fails to classify actions with large variations.

        Some deep learning methods have been proposed recently to extract keyframes \cite{Unpaired:rochan2019video,DeepKF2_kar2017adascan,DeepKF_mahasseni2017unsupervised}. A generative deep network architecture is proposed based on variational recurrent auto-encoders and generative adversarial networks (GAN) to extract keyframes  \cite{DeepKF_mahasseni2017unsupervised}. The keyframes are selected such that the video can be reproduced using the keyframes. Another keyframe selection deep network architecture is proposed which selects important frames in a single pass from a video sequence \cite{DeepKF2_kar2017adascan}. This architecture can be used with any deep CNN based models. But, both methods are unable to select a small number of keyframes to represent a video. They achieve state-of-the-art recognition accuracies by selecting almost 50\% of the frames from a video sequence, which is not better than flipping a coin. On the other hand, our method uses only 25\%-30\% of the frames from a video sequence to represent an action and achieves comparable or better results.
 

\section{Proposed Approach}
   The proposed method of online keyframe extraction module (abbreviated as OKFEM) is built upon a locally-consistent deformable convolution (LCDC) with an underlying ResNet CNN. It has been shown that LCDC has succeeded in the task of object detection and semantic segmentation \cite{LCDC:mac2019learning}. The deformable convolution has an adaptive receptive field which can capture motion naturally. In general, we observe that deformable convolution aggregates important pixels because the network has the flexibility to select samples from which each convolution uses. In a way, the adaptive receptive fields are performing some form of keypoint detection. In our method, we leverage the use of LCDC and thus the receptive fields of each frame to detect keypoints between frames in a video sequence. Not only the change in deformation of motion is detected between frames but also a pixel-wise trainable threshold kernel $TH$ is learned to select the keyframes. 
   Our method is a simple yet effective module because it only accounts for the change in motion among the neighboring frames in an arbitrary video sequence based on the previously learned threshold only. Thus it is well suited for any video sequences without having to focus on only different action classes.
   Moreover, our proposed OKFEM is an online module which can process frames sequentially and does not require future information to select keyframes. 
   We also propose a new loss function that balances the tradeoff between the accuracy and the number of keyframes selected. 
   
   For action recognition, extracted keyframes are used as input to the traditional action recognition models instead of using the full video sequence as input. To improve the performance of the traditional recognition model, we also propose a plugin module and a novel ITTS approach to incorporate the semantic word vector along with keyframes features as input to the model.   
   
   In this section we discuss our proposed module in detail with the proposed loss function. Then we show a generic model of action recognition using our proposed approach.
    
   \subsection{Online Keyframe extraction module (OKFEM)}
        \begin{figure}
        \centering
        \includegraphics[width=0.9\linewidth]{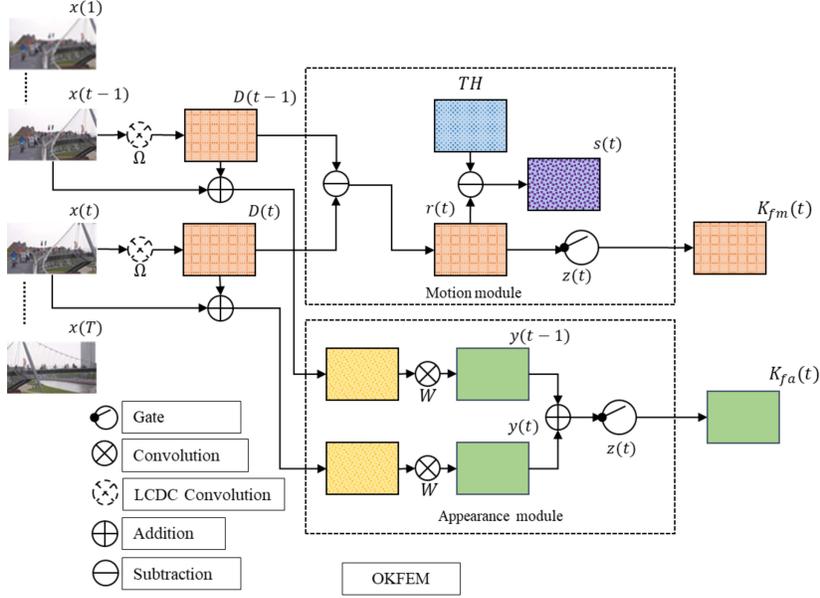}
        \caption{High level representation of the proposed online keyframe extraction module (OKFEM).}
        \label{fig:KFEM}
        \end{figure}
        The proposed OKFEM takes as input one frame of a video sequence at a time. Let $x$ be the input video sequence. At time $t$, $x(t)$ frame is fed into the OKFEM as shown in Fig.~\ref{fig:KFEM}. Then it performs LCDC using kernel $\Omega$ to produce the corresponding adaptive receptive field for frame $x(t)$. Let $D(t)$ be the adaptive receptive field for the input frame $x(t)$. Similarly, at $(t-1)_{th}$ time, the adaptive receptive field for frame $x(t-1)$ is $D(t-1)$.  Then the temporal difference of the receptive fields is calculated to extract motion information, which is denoted as:
        \begin{equation}
            \label{eqn: diffmotion}
            r(t) = D(t)-D(t-1).
        \end{equation}
        
        Using LCDC, this module can effectively model motion and appearance at time $t$. Both the motion and appearance modules are discussed below.
        \subsubsection{Motion sub-module:} 
            The difference in the receptive fields at time $t$ is $r(t)$ (using Eq. \ref{eqn: diffmotion}), which has a behavior similar to motion information produced by optical flow \cite{LCDC:mac2019learning}. Thus $r(t)$ captures the change in motion at time $t$, which is of size $H\times W$. Then we apply a learnable threshold kernel called $TH$ which is of the same size as $r$ and is updated similar to other kernels. The goal of $TH$ is to select appropriate frames as keyframes. Shown below, the difference in motion $r(t)$ is compared with $TH$ to decide whether or not the $t_{th}$ frame is a keyframe:
            \begin{equation}
                \label{eqn:compTH}
                S(t) = \sum_{h,w \in H,W} s(t)_{(h,w)} = r(t) - TH.
            \end{equation}
            Here, $S(t)$ is the score of frame $t$, which can be either positive or negative. The corresponding gate function $Z(t)$ marks the $t_{th}$ frame as a keyframe if $S(t)$ is positive and is denoted as:
            \begin{equation}
                \label{eqn:getefunction}
                Z(t) = 
                    \begin{cases}
                        1, S(t) >0,\\
                        0, otherwise.
                    \end{cases}
            \end{equation}
            Finally, the motion feature of frame $t$ is selected as key motion feature $K_{fm}(t)$ if $Z(t)=1$ and can be denoted as:
            \begin{equation}
                \label{eqn:keymotion}
                K_{fm}(t) = 
                    \begin{cases}
                        r(t), Z(t) =1,\\
                        0, otherwise.
                    \end{cases}
            \end{equation}
        \subsubsection{Appearance sub-module:}
            As shown in Fig.~\ref{fig:KFEM}, the original frame $t$ is added with the deformed receptive field $D(t)$ and the result is convolved with kernel $W$ to produce the appearance information $y(t)$ for the $t_{th}$ frame. Then based on the decision of the gate function in Eq. \ref{eqn:getefunction}, $ y(t)$ and $y(t-1)$ are added to form the key appearance feature $K_{fa}(t)$ for the  $t_{th}$ frame and can be denoted as:
            \begin{equation}
                \label{eqn:keyappearance}
                K_{fa}(t) = 
                    \begin{cases}
                        y(t)+y(t-1), Z(t) =1,\\
                        0, otherwise.
                    \end{cases}
            \end{equation}
            
    \subsection{Objective function}
        The proposed OKFEM is trained to minimize the keyframe selection loss as well as to maximize the keyframe score $S(K_f)$ (see Eq. \ref{eqn:compTH}) using the following objective function as shown in Eq. \ref{eqn:loss}. 
        \begin{equation}
        \label{eqn:loss}
            loss = min \Bigg( \alpha \sum_{f=1}^F (1-y) - \beta \sum_{f=1}^F S(K_f) \Bigg ).
        \end{equation}
        Here, $y=\begin{cases}0, f \in Y,\\ 1, f \notin Y \end{cases}$, where $Y$ is the ground truth keyframe. $\alpha$ and $\beta$ are the fine tune parameters to balance between minimizing the keyframe selection loss and selecting the higher score keyframes. In particular, $\alpha$ controls the keyframe selection accuracy and $\beta$ how many keyframes should be selected. We perform empirical study to determine the appropriate values for both parameters.
        
    \subsection{Action Recognition Model}
        \begin{figure}[ht]
        \centering
        \includegraphics[width=0.9\linewidth]{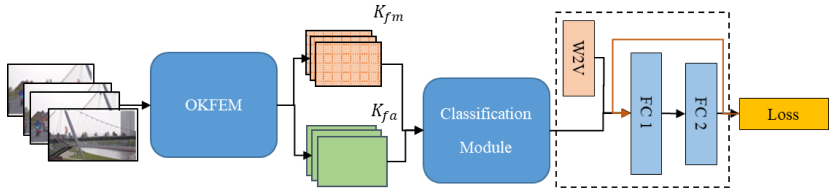}
        \caption{An overview of an action recognition model. First, keyframes are extracted using the proposed OKFEM and then use the extracted keyframes as input to an action recognition model, e.g. I3D. A plugin module (in dotted rectangle) is added at the last layer of the recognition model to adopt the W2V features.}
        \label{fig:MethodArchitecture}
        \end{figure}
        The proposed OKFEM can be used to extract keyframes from video sequences for action recognition. 
        Because of the similar objective of keyframes and key shots in summarization, our module can be trained using the video summarization datasets and test on the action recognition datasets to extract keyframes for action recognition. Thus our method can work with the transfer learning environment. The extracted keyframes from our proposed OKFEM can be used as input to either a two-stream action classification model or a 3D CNN model. An overview of the high level representation of action recognition model is shown in Fig.~\ref{fig:MethodArchitecture}.
        In particular, for the two-stream model, the extracted motion features ($K_{fm}$) and the appearance features ($K_{fa}$) of the keyframes can be used separately for the two streams. We concatenate the motion features with the appearance features to produce frame level features for the 3D CNN model.
        
\begin{algorithm}[H]
    \SetAlgoLined
    \SetKwInOut{Input}{Input}
    \SetKwInOut{Output}{Output}
    \SetKwFor{For}{for }{ do }{end}
    \Input{Keyframes of video instances $KF \in X$; Embedding word vectors of all the classes $W2V \in C$; maximum iteration $t$.}
    
    \For{each video sequence keyframes $kf \in x^{c}$ and $w2v \in W2V^{c}$}{
        \For{each iteration $i \in t$}{
            Train the classification model in Fig. \ref{fig:MethodArchitecture} using input $kf$ and $w2v$\;
            Update $w2v$ with the output of FC2 in Fig. \ref{fig:MethodArchitecture}\;
            Calculate the classification loss and update the train parameters\;
            Record predicted label $L_i$ for iteration $i$\;
            \eIf{$L_i$ is equal to $L_{i-1}$ and $L_{i-2}$ }{
               Stop iteration\;
           }{
               Continue iteration\;
          }
        }
    }
    \caption{Iterative training strategy (ITTS-Train)}
    \label{alg:training_strategy}
\end{algorithm}
        We also propose a plugin module which is a combination of two layers of fully connected networks with a feedback connection back to the first layer's input (as shown in Fig. \ref{fig:MethodArchitecture} (inside the dotted rectangle)). This module takes the combination of visual features of a particular video and W2V of its corresponding class label as input and outputs a feature vector of size equal to the original size of the W2V. We called this new feature vector as the refined W2V and is fed back to the module as well as is used to calculate the classification loss. 
        As the module has a feedback connection, we propose the ITTS algorithms. This helps us to achieve a more stable and generalized trained model.

        During training, we pass the keyframes of a particular video along with the W2V of the corresponding class label. Then in the second pass, the parameters are updated based on the current classification loss. The updated W2V is used as a new input to the plugin module with the same input video KFs features. We iterate the new W2V with the same video KFs until we get the same predicted label for three consecutive times or iterate for a maximum of 10 times and update the parameters each time based on the loss. The pseudo code for the training process is shown in Algorithm \ref{alg:training_strategy}. We observe that more than 96\% of videos achieve a consistent classification label before reaching the maximum number of iterations, which shows the significance of updating the W2V in getting consistent and more accurate classification accuracy.
        
\begin{algorithm}[H]
    \SetAlgoLined
    \SetKwInOut{Input}{Input}
    \SetKwInOut{Output}{Output}
    \SetKwFor{For}{for }{ do }{end}
    \Input{Keyframes of video instances $KF \in X$; Embedding word vectors of all the classes $W2V \in C$; maximum iteration $t$.}
    
    \For{each video sequence keyframes $kf \in x$}{
        \For{each $w2v^c \in W2V^{C}$}{
        
            \For{each iteration $i \in t$}{
                Predict the class label using input $kf$ and $w2v^c$\;
                
                Record predicted label $L_i$ for iteration $i$\;
                
                Update $w2v^c$ with the output of FC2 in Fig. \ref{fig:MethodArchitecture}\;

                \eIf{$L_i$ is equal to $L_{i-1}$ and $L_{i-2}$ }{
                   Stop iteration\;
               }{
                   Continue iteration\;
              }
            }
            Record predicted label $L_c$ with probability score $P_c$ for $w2v^c$\; 
        }
        Calculate predicted label $L_{kf}$ for $x$ using $P_c \in C$ and $L_c \in C$ \;
    }
    \caption{Iterative testing strategy (ITTS-Test)}
    \label{alg:testing_strategy}
\end{algorithm}
        During testing, with each test video instance, we fuse the W2Vs from all the classes one at a time and record the probability score. The one with the highest score is selected as the predicted class. For each W2V, we also perform iteration with the updated W2V similar to that during training assuming the hypothesis of predicting the same class label for three consecutive times achieves more stable and accurate results. The pseudo code for the training process is shown in Algorithm \ref{alg:testing_strategy}.
   
\section{Experiments}
    \subsection{Implementation details}
        The proposed method is implemented using ResNet50 with deformable convolution as backbone as in \cite{LCDC:mac2019learning}. We resize the frames to 224 $\times$ 224 and use full frame as input to our module. We use the common momentum optimizer (with momentum = 0.9).The learning rate was initialized as $10^{-4}$ with a decay rate of $0.96$ for every 10 epochs. A total of 30 epochs were used for training. We use the video summarization datasets to train/test our module. For action recognition, we use the module trained using the summarization datasets and applied it to the action recognition datasets to extract keyframes. We train a classification model using the extracted keyframes of the training set of an action recognition dataset. Then, we test the model using the keyframes of the test or validation set of the same dataset.
        
        For action recognition, we use two popular state-of-the-art models as baseline classification models. First, the I3D model \cite{carreira2017quo}, which uses an RGB and a flow stream as the two-stream model and the other is the popular 3D ResNet architecture with two variations: ResNet18 and ResNet152 as the 3D CNN model. For the 3D CNN models, we concatenate the motion and appearance features before feeding them into the network. In the plugin module, we use a output size of 450 and 300 for the FC1 and FC2, respectively. 
        
    \subsection{Datasets and Evaluation Metrics}
        \paragraph{Datasets} 
            We evaluate our method for both video summarization and action recognition. For video summarization, we train/test our proposed OKFEM using two well-known video summarization datasets, namely   SumMe \cite{SumMe:gygli2014creating} and TVSum \cite{TVSum:song2015tvsum}. We leverage the use of YouTube and OVP datasets for training only following the practice of previous works \cite{Susinet:koutras2019susinet,Unpaired:rochan2019video}. We evaluate our method on SumMe and TVSum datasets following prior work \cite{Susinet:koutras2019susinet,Unpaired:rochan2019video}. The SumMe dataset has 25 videos and the TVSum dataset has 50 videos. For evaluation, we use 20\% videos from SumMe or TVSum dataset for testing and the rest of the data from SumMe, TVSum, YouTube, and OVP datasets are used for training, which is called \textit{Augmented} setting. Moreover, we also use a \textit{transfer} setting by which the test dataset is completely isolated from training and only the other three datasets are used for training.
            
            For action recognition, we use three well-known benchmark action recognition datasets: Kinetics-400 \cite{Kinetic400:kay2017kinetics}, Something-Something-V1 \cite{SSv1dataset:goyal2017something}, and HMDB51 \cite{Dataset_hmdb51}.
            
            Kinetics-400 consists of 400 actions. Videos are collected from YouTube and most of them are around 10 seconds long. The training set consists of around 240,000 videos and the validation set of around 19,800 videos. It is a well-balanced dataset with large scale diversity. We report the results on the validation set.
           
           Something-Something-V1 consists of 174 actions. The dataset contains around 110,000 videos. The video duration typically spans from 2 to 6 seconds. This dataset is well known due to its diversity of actions that requires temporal information to recognize. The results are reported on the standard test set.
           
           HMDB51 dataset contains 6849 video sequences divided into 51 action categories. Video sequences are extracted from commercial movies and YouTube. The videos are of different resolutions and frame rates. This dataset has three standard train and test splits. We report the average accuracy of the standard test splits.

      \paragraph{Evaluation metrics}
          For video summarization, we evaluate our proposed module using key-shot based protocol as suggested in \cite{keyshotProtocol:zhang2016video,Unpaired:rochan2019video,Susinet:koutras2019susinet}. We convert the extracted keyframes using our proposed method to key-shots following the approach in \cite{keyshotProtocol:zhang2016video} which uses the KTS algorithm \cite{KTS:potapov2014category} to temporally segment the videos into segments. Segments are ranked according to the ratio between the number of keyframes in a segment and the length of the segment. Then we choose the segments that are at most 15\% of the original video. 
          The TVSum dataset provides frame-level importance scores which we also convert to key-shots as done by \cite{keyshotProtocol:zhang2016video,Unpaired:rochan2019video} for evaluation. The SumMe dataset has keyshot-based ground truth annotation, which is used directly for evaluation.
          Following the evaluation protocol of the datasets and prior work \cite{keyshotProtocol:zhang2016video,Unpaired:rochan2019video}, we compute the F-score of the predicted key-shots and the ground truth. 
          
          For action recognition, 
          the classification models are evaluated based on the recognition accuracy of the  validation or the test sets of the datasets.

    \subsection{Analysis of the Module}
        In this study we first validate the proposed OKFEM module using video summarization datasets. For evaluation, we use only SumMe and TVSum datasets. Here, we show the importance of the $\alpha$ and $\beta$ parameters in the loss function. Fine tuning these parameters have improved the results significantly, which suggests the importance of defining the loss function with these parameters. Table  \ref{table:alphabeta} shows the results of using different values of $\alpha$ and $\beta$ and the corresponding F-score for the SumMe and TVSum datasets. 
        \setlength{\tabcolsep}{4pt}
        \begin{table}
        \begin{center}
        \caption{F-score (in \%) for the SumMe and TVSum datasets using different values of $\alpha$ and $\beta$ in the loss function.}
        \label{table:alphabeta}
        \begin{tabular}{|c|c|c|c|}
        \hline
        $\alpha$ & $\beta$ & SumMe & TVSum\\
        \hline
        \hline
        0.2	& 0.8 & 46.3\% & 59.2\%\\
        0.4	& 0.6 & 47.4\% & 61.3\%\\
        0.5 & 0.5 & 47.2\% & 61.5\%\\
        0.5 & 0.45 & 47.3\% & 61.4\%\\
        0.55 & 0.45 & 47.3\% & 61.4\%\\
        \textbf{0.6}	& \textbf{0.42} & \textbf{47.9\%} &	\textbf{62.4\%}\\
        0.6	& 0.4 & 47.7\% & 61.7\%\\
        0.62 & 0.42	& 47.8\% & 62.1\%\\
        0.64 & 0.42	& 47.6\% & 61.7\%\\
        0.8	 & 0.2	& 44.8\% & 57.5\%\\
        \hline
        \end{tabular}
        \end{center}
        \end{table}

        Based on the experimental results using different values of $\alpha$ and $\beta$ as shown in Table \ref{table:alphabeta}, it is clear that carefully adjusting the parameters can significantly improve the F-score (almost 3\% for SumMe and 5\% for TVSum) for both datasets. The OKFEM module achieves the best results using $\alpha=0.6$ and $\beta=0.42$. It is noteworthy that our proposed method on average extracts at most 25-30\% frames as keyframes which is very low compared to other state-of-the-art methods \cite{DeepKF2_kar2017adascan,DeepKF_mahasseni2017unsupervised}. We use these parameter values for the rest of the experiments video summarization methods.
        
    \subsection{Ablation Study}
        We perform an ablation study to observe the effectiveness of the proposed OKFEM module in keyframe extraction in terms of action recognition. In this experiment, we use three sets of data from a dataset. First, our extracted keyframes are used as input to the classification model. Second, as mentioned before our method extracts on average of at most 30\% of the video sequence as keyframes. Thus for this study, we randomly select 30\% of the frames and use them as keyframes to the classification model to evaluate the importance of our extracted keyframes. For this random frames we extracted the motion and appearance features without applying the threshold $TH$. It is noteworthy that the randomly chosen frames may contain the extracted keyframes using our OKFEM. Third, we also prepare another set of 30\% of randomly selected frames that excludes the extracted keyframes. Now we perform classification on these three sets of data to see the effectiveness of our proposed OKFEM module for extraction. For this experiment, we use the Kinetics-400 and Something-Something-V1 datasets. 
         \setlength{\tabcolsep}{4pt}
        \begin{table}[ht]
        \begin{center}
        \caption{Classification accuracy in \% on the Kinetics-400 and Something-Something-V1 datasets using keyframes extracted from OKFEM (KF Set), randomly selected 30\% frames which may include extracted keyframes (R Set1), and randomly selected 30\% frames which exclude extracted keyframes (R Set2).}
        \label{table:ablationstudy}
        \begin{tabular}{|c||c|c|c||c|c|c|}
        \hline
        
        \multirow{2}{*}{Classification Model} & \multicolumn{3}{c||}{Something-Something-V1}& \multicolumn{3}{c|}{Kinetics-400}\\
        \cline{2-7}
          & KF Set & R Set1 & R Set2 & KF Set & R Set1 & R Set2\\
        \hline
        \hline
        OKFEM+I3D	& \textbf{43.8} & 39.5 & 37.8 & \textbf{75.8} & 69.2 & 53.4\\
        OKFEM+ResNet18 & \textbf{45.7}	& 42.8 & 40.4 &	\textbf{76.3} & 72.5	& 49.8\\
        OKFEM+ResNet152 & \textbf{52.3} & 46.4 & 41.2 & \textbf{77.5} & 73.4 & 56.7\\
        \hline
        \end{tabular}
        \end{center}
        \end{table}
        
        From Table \ref{table:ablationstudy}, we see that the classification accuracy using our extracted keyframes clearly outperforms the other two random sets for both datasets. Moreover, we see that sets (R Set2) that exclude the keyframes perform worst in the classification. This is true for all the three classification models. These results show the importance of our OKFEM module for keyframe extraction. We also observe from Table \ref{table:ablationstudy} that there is a  significant improvement in accuracy (more than 10\% for Something-Something-V1 and 20\% for Kinetics-400) from R Set2 to KF Set frames because of carefully chosen keyframes using our proposed OKFEM method.   
        
        \begin{table}[ht]
        \begin{center}
        \caption{Classification accuracy (in \%) comparison with the baseline 3D CNN methods on the Kinetics-400 and  Something-Something-V1 datasets. The best results are marked as \textcolor{red}{\textit{red}} for each method.}
        \label{table:baselineAccuracyComp}
        \begin{tabular}{|c|c|c|c|}
        \hline
        \multirow{2}{*}{Method} &  Something-& \multirow{2}{*}{Kinetics-400}\\
           & Something-V1 &\\
        \hline
        \hline
        I3D \cite{carreira2017quo} & 41.6 & 74.2\\
        ResNet18 & 37.6 & 55.4\\
        ResNet152 & 40.8 & 57.2\\
        \hline
        OKFEM+I3D & {43.8} & {75.8}\\
        OKFEM+ResNet18  & {45.7} & {76.3}\\
        OKFEM+ResNet152  & {52.3} & {77.5}\\
        \hline
        OKFEM+W2V+I3D & \textcolor{red}{48.2} & \textcolor{red}{82.4}\\
        OKFEM+W2V+ResNet18  & \textcolor{red}{49.4} & \textcolor{red}{82.3}\\
        OKFEM+W2V+ResNet152  & \textcolor{red}{54.8} & \textcolor{red}{83.6}\\
        \hline
        \end{tabular}
        \end{center}
        \end{table}
        We also compare our proposed classification models with that of the  baseline classification models. The baseline models use the full video sequence as input whereas only keyframes are used as input (for both training and testing) to the corresponding classification models. Table~\ref{table:baselineAccuracyComp} shows the recognition accuracies of the corresponding methods for the two datasets. For the I3D method, using our extracted keyframes outperforms the original I3D for both datasets. Moreover, the ResNet models with our OKFEM outperforms the original ResNet models using the full video sequence by a large margin for both datasets. In particular, OKFEM+ResNet18 achieves a performance gain of 8.1\% for the Something-Something-V1 dataset and 20.9\% for the Kinetics-400 dataset. Similar performance gain is also observed for the ResNet152 model, which clearly shows the importance of our proposed keyframe extraction method. 
        Furthermore, applying our plugin module with W2V as an additional input and using the proposed ITTS algorithms, the proposed recognition approach outperforms all the above baseline categories in Table \ref{table:baselineAccuracyComp}. These results suggest the significance of the proposed plugin module along with the new ITTS approach. 
        
    \subsection{Comparison with the State-of-the-Art}
        We compare our proposed method of OKFEM with the state-of-the-art methods in terms of both video summarization and keyframe-based action recognition. 
        
        \textbf{Video summarization:}
            For video summarization, we compare our proposed module with the state-of-the-art video summarization methods. We compare using both augmented and transfer settings of the SumMe dataset and the TVSum dataset. For both datasets, our approach achieves competitive results in terms of F-score (see Table \ref{table:vidSummaraizationComp}). 
            \setlength{\tabcolsep}{4pt}
            \begin{table}[ht]
            \begin{center}
            \caption{Comparison of the proposed OKFEM module with the state-of-the-art methods in terms of F-score (in \%) (in the augmented and transfer setting scenarios) for the SumMe and TVSum datasets. The best, second and third best results are marked as \textcolor{red}{\textit{red}}, \textcolor{blue}{\textit{blue}} and \textcolor{green}{\textit{green}}, respectively.}
            \label{table:vidSummaraizationComp}
            \begin{tabular}{|c||c|c||c|c|}
            \hline
            \multirow{2}{*}{Method} & 
            \multicolumn{2}{c||}{SumMe} & 
            \multicolumn{2}{c|}{TVSum}\\
            \cline{2-5}
                & Augmented & Transfer & Augmented & Transfer \\
            \hline
            \hline
            Zhang \emph{et al.} \cite{keyshotProtocol:zhang2016video} & 42.9 & \textcolor{green}{41.8} &  59.6 & \textcolor{blue}{58.7}\\
            Mahasseni \emph{et al.} \cite{DeepKF_mahasseni2017unsupervised} & 43.6 & - &  61.2 & -\\
            Zhao \emph{et al.} \cite{43zhao2017hierarchical} & 43.6 & - & 61.5 & -\\
            Zhou \emph{et al.} \cite{47zhou2018deep} & 43.9 & - & 59.8 & -\\
            Zhang \emph{et al.} \cite{40zhang2018retrospective} & 44.1 & - & \textcolor{red}{63.9} & -\\
            SUSiNet \cite{Susinet:koutras2019susinet} & 41.1 & - & 59.2 & -\\
            UnpairedVSN \cite{Unpaired:rochan2019video}	& 47.5 & - & 55.6 & - \\
            VASNet \cite{VASNET_fajtl2018summarizing} & \textcolor{blue}{51.0} & - & \textcolor{green}{62.3} & -\\
            SUM-FCN \cite{SUM_FCN_rochan2018video} & \textcolor{red}{51.1} & \textcolor{blue}{44.1} & 59.2 & \textcolor{green}{58.2}\\
            \hline
            OKFEM &	\textcolor{green}{47.9} &\textcolor{red}{45.6} &	\textcolor{blue}{62.4} & \textcolor{red}{60.9}\\
    
            \hline
            \end{tabular}
            \end{center}
            \end{table} 
            For the SumMe dataset, our method is comparable to the state-of-the-art methods in augmented environment where our method is a simple yet online approach to extract keyframes. Our module achieves comparable results to the best method on the TVSum dataset, and is the second best among the state-of-the-art methods in the augmented environment. In the transfer environment, our method outperforms all the state-of-the-art methods by a significant margin, which inspires us to apply our proposed OKFEM module to extract keyframes for the action datasets.
            
        \textbf{Action recognition:}
              Our extracted keyframes are applied to two well-known 3D CNN methods called I3D and two variations of ResNet. The experimental results are shown in Table \ref{table:accuracyComp}. The classification accuracies of state-of-the-art methods reported in literature are compared with that of I3D and ResNets that use keyframes as input. Even with using only the keyframes in I3D and ResNets, the recognition accuracies are closely comparable to that of the state-of-the-art methods. Moreover, including W2V as input along with keyframes using the plugin module and applying the novel ITTS algorithms, the results outperform all the state-of-the-art methods for the three datasets. 
            \begin{table}[ht]
            \begin{center}
            \caption{Classification accuracy (in \%) comparison with the state-of-the-art methods on the Kinetics-400,  Something-Something-V1, and HMDB51 datasets. The best, second and third best results are marked as \textcolor{red}{\textit{red}}, \textcolor{blue}{\textit{blue}} and \textcolor{green}{\textit{green}}, respectively.}
            \label{table:accuracyComp}
            \begin{tabular}{|c|c|c|c|}
            \hline
            Method & HMDB51 & Something-& Kinetics-400\\
             &  & Something-V1 &\\
            \hline
            \hline
            I3D \cite{carreira2017quo} & \textcolor{green}{80.2} & 41.6 & 74.2\\
            R(2+1)D \cite{closer_tran2018closer} & 78.7 & - & 75.4\\
            Non-local NN \cite{nonlocalNN:wang2018non} & - & - & {77.7}\\
            Brais \emph{et al.} \cite{FilerBank:martinez2019action}+ResNet18 &  - & 45.0 & -\\
            Brais \emph{et al.} \cite{FilerBank:martinez2019action} +ResNet152 & - & \textcolor{blue}{53.4} & {78.8}\\

            \hline \hline
            OKFEM+I3D & \textcolor{blue}{83.8} & 43.8 & 75.8\\
            OKFEM+ResNet18 & - & 45.7 & 76.3\\
            OKFEM+ResNet152 & - & \textcolor{green}{52.3} & {77.5}\\
            \hline
            OKFEM+W2V+I3D & \textcolor{red}{87.9} & {48.2} & \textcolor{blue}{82.4}\\
        OKFEM+W2V+ResNet18 & - & {49.4} & \textcolor{green}{82.3}\\
        OKFEM+W2V+ResNet152 & - & \textcolor{red}{54.8} & \textcolor{red}{83.6}\\
        \hline
            \end{tabular}
            \end{center}
            \end{table}
            
            For the HMDB51 dataset, OKFEM+I3D outperforms the original I3D method by a significant margin (by 3.6\%) which demonstrates the effectiveness of our proposed OKFEM. Moreover, the results show that unnecessary frames may hinder recognition accuracy. Thus, useful keyframe extraction methods are helpful for better action recognition. Furthermore, applying the plugin module with the ITTS algorithms our OKFEM+W2V+I3D achieves the best accuracy of 87.9\% which is even better than OKFEM+I3D by 4.1\%. This result demonstrates the effectiveness of our plugin module with the proposed ITTS algorithms. 
            
            Our OKFEM+Baseline methods closely compete with state-of-the-art methods for the Something-Something-V1 and Kinetics-400 datasets. It is noteworthy that all the state-of-the-art methods use full video sequences as input for action recognition whereas using our extracted keyframes, the same methods achieve almost similar results. Indeed, our OKFEM+ResNet152 is the second-best for Something-Something-V1 compared to the state-of-the-art methods and closely competes with the Non-local NN method for the second position for the Kinetics-400 datasets. 
            Thus, it again supports the importance of keyframes and the effectiveness of our OKFEM module for the extraction of keyframes. Moreover, using W2V, the proposed OKFEM+W2V+ResNet152 outperforms all the state-of-the-art methods and achieves the best results for the Something-Something-V1 and Kinetics-400 datasets.
        
        \subsection{Qualitative Analysis}
             For the qualitative analysis, we use two sample videos from the SumMe dataset. We compare our OKFEM with that of the state-of-the-art UnpairedVSN \cite{Unpaired:rochan2019video} method in terms of video summarization. 
            \begin{figure}[ht]
            \centering
            \includegraphics[width=0.85\linewidth]{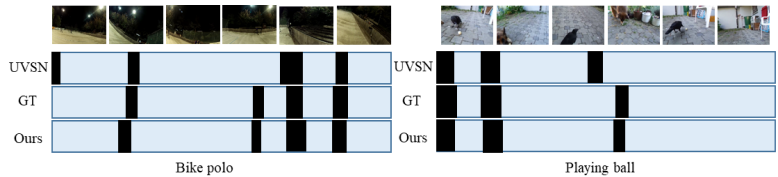}
            \caption{Two sample videos from the SumMe dataset. The bars represent the video timeline with the summaries created by each method (UnpairedVSN \cite{Unpaired:rochan2019video}(UVSN) and OKFEM) along with the ground truth. The black blocks represent sequences of frames selected as the summary blocks.}
            \label{fig:summaryComp}
            \end{figure}
           From Fig.~\ref{fig:summaryComp}, we observe that video summary segments (black blocks) generated by our approach closely match with that of the ground truth for both video sequences. These results demonstrate the effectiveness of our OKFEM for video summarization. 
            
            
            \begin{figure}
            \centering
            \includegraphics[width=0.9\linewidth]{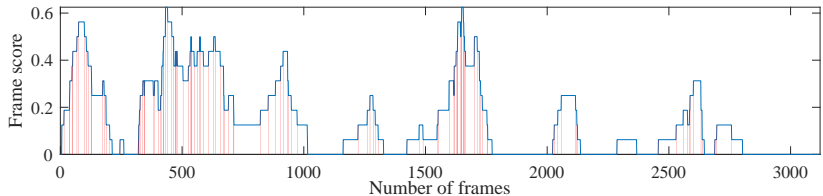}
            \caption{Ground truth frame label score (in \textcolor{blue}{blue} curve) of \emph{Playing ball} video of SumMe dataset \cite{SumMe:gygli2014creating} along with the selected keyframes (in \textcolor{red}{red} lines) of the same video using our OKFEM.}
            \label{fig:keyframeComp_playingball}
            \end{figure}
            We also demonstrate the effectiveness of our proposed OKFEM in keyframe selection by showing the selected keyframes along with the ground truth frame level score of a video sequence as shown in Fig.~
            \ref{fig:keyframeComp_playingball}. The blue curve in the figure represents the frame level ground truth score. The higher the value in score the more important the corresponding keyframe. In the figure, red lines represent the selected keyframes by our OKFEM. From the figure, we observe that most of the frames selected as keyframes have high frame level scores. This suggests the appropriateness of our method in keyframe selection. 

\section{Conclusions}
    A novel online learnable keyframe extraction module is proposed to extract meaningful frames that are sufficient to represent a video sequence. In particular, we use a learnable threshold to select keyframes. The proposed module extracts keyframes in an online fashion by examining frames in a video sequence one at a time and select keyframes based on processed frames only. To the best of our knowledge, we are the first to propose such an online keyframe extraction module.  
    The experimental results demonstrate the effectiveness of the proposed module for keyframe extraction which extracts less than 30\% of the frames as keyframes. Moreover, the proposed OKFEM performs well in the transfer learning environment. Thus, not only the OKFEM can be used for video summarization but also for action classification. The proposed loss function along with appropriate parameter values can improve the accuracy of keyframe selection. Therefore, our proposed module combined with the proposed loss function is a promising tool to extract keyframe in an online fashion. Moreover, the traditional action classification model using OKFEM can compete with state-of-the-art methods. Finally, applying our proposed plugin module with the novel ITTS algorithms outperforms all other methods in action classification and demonstrates the significance of it.
    In the future, we will apply the extracted keyframes for online real time action recognition. Moreover, we will improve the OKFEM to select only one set of keyframes for repetitive actions.


\bibliography{references}

\end{document}